\documentclass{article}
\usepackage{algorithm}
\usepackage{algpseudocode}
\usepackage[dblblindworkshop,final]{{neurips_2026}}
\workshoptitle{{SLMs for Agentic Systems, Paris, France, 2026}}

\usepackage[utf8]{inputenc} 
\usepackage[T1]{fontenc}    
\usepackage{hyperref}       
\usepackage{url}            
\usepackage{amsmath,amssymb,booktabs,multirow,array,graphicx,url,xcolor}
\usepackage{tikz}
\usetikzlibrary{arrows.meta,positioning,shapes.geometric,fit,calc}
\usepackage{booktabs}       
\usepackage{amsfonts}       
\usepackage{nicefrac}       
\usepackage[
  format=plain,
  labelsep=colon,
  justification=justified,
]{caption}
\usepackage{microtype}      
\usepackage{xcolor}         
\newcommand{\method}{\textsc{StepGate}}

\title{Do I Need the Cloud? Uncertainty-Aware Step-Level Handoff for Small Language Model Agents}
\workshoptitle{{SLMs for Agentic Systems, Paris, France, 2026}}

\author{%
  Abolfazl Younesi \\
  Department of Computer Engineering \\
  Sharif University of Technology \\
  Tehran, Iran \\
  \texttt{abolfazl.younesi@ieee.org} \\
}

\begin{document}

\maketitle

\begin{abstract}
Small language models (SLMs) are attractive as local agent controllers because they reduce remote inference, latency, and deployment footprint, yet structured tool errors can cause an agent step to fail. Existing routers typically select a model once per query. However, agents expose sequential decision points whose difficulty dynamically changes based on intermediate observations. We propose \method, an uncertainty-aware handoff framework that scores each local SLM action and selectively escalates challenging steps to a stronger model. On a 52-task held-out single-step BFCL-derived test split, the Qwen2.5-1.5B/7B pair attains 82.7\% task success with 30.8\% escalation, versus 67.3\% local-only and 75.4\% random escalation  {(which uses 33.8\% escalation)}. In a separate multi-turn evaluation, \method{} achieves 69.0\% trajectory success and 84.0\% action success using only 30.0\% cloud actions, compared with 48.0\%/70.5\% local-only, 60.0\%/78.2\% random escalation, and 57.0\%/77.1\% query-level routing (strong-only achieves 82.0\% trajectory success at 100\% cloud actions).  {These results suggest that step-level escalation recovers a large share of the performance gap to the stronger Qwen2.5-7B backend at a matched cloud-action rate while transmitting fewer tokens remotely. However, our evaluation is limited to one model family, a single stronger backend, and scripted tasks. Furthermore, the test sets are small, multi-turn comparisons rely on paired intervals and statistical tests, and our risk tiers serve as research annotations rather than formal safety guarantees.}
\end{abstract}

\section{Introduction}
Agentic systems repeatedly invoke language models for narrow decisions: selecting a tool, filling arguments, interpreting an observation, or choosing the next action. This execution pattern strengthens the case for SLMs: many calls are repetitive and specialized, and heterogeneous systems can reserve large models for the minority of steps that truly require them \citep{belcak2025slm}. The difficulty is reliability. Function calling is structurally unforgiving: a nearly correct natural-language answer may still be useful, whereas one wrong function name, entity identifier, enum value, or destructive argument can fail an entire trajectory. BFCL consequently evaluates serial/parallel calls, abstention, stateful multi-turn behavior, and increasingly holistic agentic behavior rather than text quality alone \citep{patil2025bfcl}.
Model routing is a natural response to heterogeneous capabilities. FrugalGPT, RouteLLM, Hybrid LLM, AutoMix, and related cascades show that selectively invoking stronger models can reduce cost while retaining quality \citep{chen2024frugalgpt,ong2025routellm,ding2024hybrid,aggarwal2024automix,yue2024cascade}. However, these systems mostly route at \emph{query level} and optimize answer quality/cost. An agent instead exposes a sequence of decisions whose difficulty can change after every tool result. A request such as ``find my meeting and move it if the room is unavailable'' may require an easy local search, a difficult disambiguation after observing multiple events, and a high-risk calendar mutation. Routing the entire request to the cloud wastes local capability. Committing the entire trajectory to an SLM compounds errors.

We therefore ask: \emph{Can an SLM predict, before execution, when its next agent action is likely to fail and use that uncertainty to escalate only the necessary step?} This question differs from generic LLM routing in three ways. First, the decision unit is an \emph{action}, not a user query. Second, uncertainty is action-structured: tool identity, arguments, schema validity, execution state, and action risk provide signals unavailable in ordinary QA. Third, the objective is multi-dimensional: cloud calls and monetary cost matter, but so do local latency/energy, end-to-end task success, and the amount of private context disclosed remotely.

\begin{figure}[t]
\centering
\resizebox{\linewidth}{!}{%
\begin{tikzpicture}[
    font=\scriptsize,
    node distance=5mm and 6mm,
    box/.style={
        draw,
        rounded corners,
        align=center,
        minimum height=7mm,
        inner sep=3pt
    },
    small/.style={box,fill=gray!7},
    strong/.style={box,fill=gray!18},
    dec/.style={
        diamond,
        draw,
        aspect=2.0,
        align=center,
        inner sep=1.5pt
    },
    arr/.style={-{Latex[length=2mm]},thick}
]
\node[small] (state)
    {Agent state $s_t$\\user goal + history + tools};
\node[small,right=of state] (slm)
    {Local SLM\\propose $a_t$};
\node[small,right=of slm] (feat)
    {Action uncertainty\\$\phi(s_t,a_t)$};
\node[dec,right=8mm of feat] (gate)
    {$\hat p_{\rm fail}$\\+ risk};
\node[small,above right=7mm and 9mm of gate] (local)
    {Execute locally};
\node[small,right=9mm of gate] (verify)
    {Local verify\\or retry};
\node[strong,below right=8mm and 9mm of gate] (cloud)
    {Strong LLM\\handoff};
\node[small,right=12mm of verify] (tool)
    {Tool / environment\\observation $o_{t+1}$};
\draw[arr] (state) -- (slm);
\draw[arr] (slm) -- (feat);
\draw[arr] (feat) -- (gate);
\draw[arr]
    (gate.north east) --
    node[above,sloped,pos=.48]{low risk}
    (local.south west);
\draw[arr]
    (gate.east) --
    node[above,pos=.52]{medium}
    (verify.west);
\draw[arr]
    (gate.south east) --
    node[below,sloped,pos=.40]{high risk}
    (cloud.north west);
\draw[arr]
    (local.east) -| (tool.north);
\draw[arr]
    (verify.east) -- (tool.west);
\draw[arr]
    (cloud.east) -| (tool.south);
\coordinate (feedbackY) at ($(cloud.south)+(0,-5mm)$);
\draw[arr,dashed]
    (tool.south)
    |- (feedbackY)
    -| node[pos=.62,below]{next step}
    (state.south);
\end{tikzpicture}%
}
\caption{\method{} routes \emph{each agent step}. The local SLM first proposes an action. A lightweight gate estimates failure probability from action-structured uncertainty and risk. Only uncertain/high-risk steps disclose the required context to a stronger model.}
\label{fig:arch}
\end{figure}
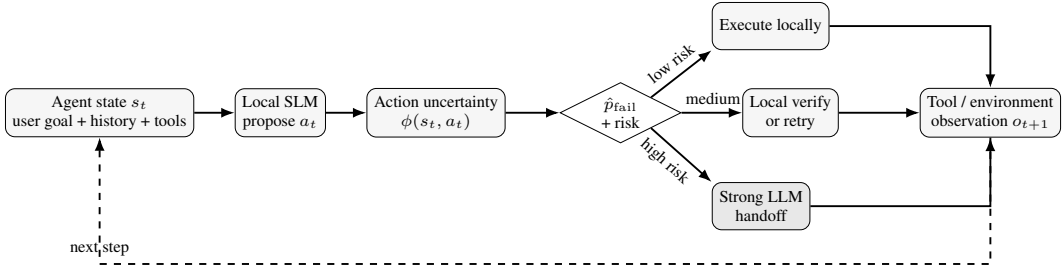

Our contributions are fourfold. \textbf{(1)} We formulate action-level SLM--LLM handoff as a constrained risk minimization problem for tool-augmented agents, accounting for task failure, cloud invocation, latency/energy, and remote-context exposure. \textbf{(2)} We design an action-aware uncertainty representation that fuses token confidence, schema and argument structure, sample consistency, and external risk tiers. \textbf{(3)} We implement a calibrated, adaptive gate that  {improves on local-only and query-level routing at matched cloud-action rates, is positive but not clearly separated from random escalation, and matches self-consistency routing with fewer cloud actions in our multi-turn setting.} \textbf{(4)} We evaluate both single-step and multi-turn agent trajectories: at a 30.0\% cloud-action budget, \method{}\footnote{\url{https://zenodo.org/records/22308389}} improves trajectory success from 48.0\% (local-only) to 69.0\% (versus 60.0\% random and 57.0\% query-level routing), with 55.0\% of successful trajectories executing via mixed local/cloud collaboration {. We report paired bootstrap intervals and McNemar tests for these comparisons and evaluate the risk features at an equal cloud budget (Sec.~\ref{sec:experiments})}.

\section{Related work}
\label{sec:related}

\paragraph{Efficient heterogeneous inference.} LLM cascades exploit heterogeneous cost and capability. FrugalGPT learns cascades over commercial models \citep{chen2024frugalgpt}. RouteLLM learns strong-versus-weak routing from preference data \citep{ong2025routellm}. A hybrid LLM predicts query difficulty and enables quality/cost control \citep{ding2024hybrid}. AutoMix combines self-verification with sequential routing \citep{aggarwal2024automix}, and mixture-of-thought cascades use answer consistency for cost-efficient reasoning \citep{yue2024cascade}. Complementary systems work reduces inference cost through memory-aware serving and resource disaggregation: vLLM improves KV-cache management \citep{kwon2023pagedattention}, FlexGen targets constrained-memory inference \citep{sheng2023flexgen}, while Splitwise and DistServe separate inference phases across heterogeneous resources \citep{patel2024splitwise,zhong2024distserve}. Beyond server-side disaggregation, collaborative edge--cloud frameworks partition LLM execution under dynamic network and workload conditions \citep{younesi2025splitwise}. These systems optimize where and how inference executes. \method{} instead decides \emph{which model should execute the next structured agent action}. Its routing unit is the state-action pair $(s_t,a_t)$ rather than the full query.

\paragraph{Uncertainty and selective prediction.} Calibration is central when confidence controls deferral: modern neural predictors can be miscalibrated \citep{guo2017calibration}, while selective prediction formalizes risk-coverage trade-offs under abstention \citep{geifman2019selectivenet}. For language models, sampling disagreement provides a practical uncertainty signal: self-consistency aggregates multiple reasoning paths \citep{wang2023selfconsistency}, SelfCheckGPT detects inconsistent generations \citep{manakul2023selfcheck}, and semantic entropy targets meaning-level uncertainty \citep{farquhar2024semantic}. Uncertainty has also guided information-seeking plans \citep{hu2024uot}. Three further lines are closely related (Appendix~\ref{app:extended-related} discusses each in depth): TokUR estimates token-level epistemic uncertainty for free-form reasoning via weight-perturbation ensembles \citep{zhang2026tokur}, motivating our simpler decode-time intrinsic features. UALA gates \emph{whether} a ReAct-style agent consults external tools \citep{han2024uala}, a decision complementary to \method's choice of \emph{which executor} runs an already-selected call; and RACER casts model selection as $\alpha$-Valid Optimal Routing with a finite-sample, distribution-free misrouting-risk guarantee \citep{hao2026racer}, which \method's empirically calibrated gate does not yet provide (Appendix~\ref{sec:limitations}). We adapt these principles to tool actions, where similar generations may differ by a single executable argument, leading to different external consequences.

\paragraph{Agents and tool use.} ReAct interleaves reasoning with environment actions \citep{yao2023react}, Toolformer studies learned API invocation \citep{schick2023toolformer}, and Gorilla targets accurate selection and invocation from large API collections \citep{patil2024gorilla}. TinyAgent demonstrates function calling with small models at the edge \citep{erdogan2024tinyagent}, while AgentDojo highlights that tool-using agents face reliability and security failures in realistic environments \citep{debenedetti2024agentdojo}. BFCL provides standardized function-calling evaluation spanning serial/parallel calls, abstention, and stateful agent behavior \citep{patil2025bfcl}. MCP supplies an interoperable JSON-RPC/tool-schema protocol \citep{mcp2025}. These works motivate reliable tool execution. Our focus is calibrated pre-execution handoff from a local SLM to a stronger model.

\begin{table}[t]
\centering\scriptsize
\caption{Positioning against representative recent work. ``Step'' means routing can change after each tool observation. ``Structured UQ'' uses tool/argument/state signals. ``Risk'' accounts for action consequence. ``Guarantee'' denotes a finite-sample/distribution-free risk-control guarantee.}
\label{tab:related}
\setlength{\tabcolsep}{2.6pt}
\begin{tabular}{lcccccc}
\toprule
Method & Small$\to$large & Step & Structured UQ & Tool risk & Agent metric & Guarantee\\
\midrule
FrugalGPT \citep{chen2024frugalgpt} & \checkmark & -- & -- & -- & -- & --\\
RouteLLM \citep{ong2025routellm} & \checkmark & -- & -- & -- & -- & --\\
Hybrid LLM \citep{ding2024hybrid} & \checkmark & -- & -- & -- & -- & --\\
AutoMix \citep{aggarwal2024automix} & \checkmark & -- & partial & -- & -- & --\\
UALA \citep{han2024uala} & -- & partial & -- & -- & -- & --\\
RACER \citep{hao2026racer} & \checkmark & -- & -- & -- & -- & \checkmark\\
BFCL \citep{patil2025bfcl} & -- & -- & -- & -- & \checkmark & --\\
\textbf{\method{} (ours)} & \checkmark & \checkmark & \checkmark & \checkmark & \checkmark & --\\
\bottomrule
\end{tabular}
\end{table}

\section{Uncertainty-aware step-level handoff}
Let an agent interact with an environment for $T$ steps. At step $t$, local model $M_s$ observes state $s_t=(x,h_t,\mathcal{T}_t)$ containing the user goal $x$, interaction history $h_t$, and available tools $\mathcal{T}_t$, then proposes structured action $a_t=(z_t,v_t)$ with tool $z_t$ and arguments $v_t$. A routing policy $\pi$ chooses $r_t\in\{L,V,C\}$: execute locally ($L$), perform a bounded local verification/retry ($V$), or escalate the step to strong cloud model $M_l$ ($C$). The environment returns observations $o_{t+1}$ and updates the state. Unlike query-level routing, $\pi$ is re-evaluated after every observation.

\paragraph{Action-aware uncertainty.} We construct $\phi_t=\phi(s_t,a_t)$ from six signal families. \emph{Token uncertainty} summarizes length-normalized negative log-likelihood and entropy over the generated function name/arguments when logits are available. \emph{Tool margin} measures separation between the chosen tool and plausible alternatives, using normalized log probability or a lightweight scorer. \emph{Schema signals} include parse validity, required-field coverage, enum/type constraints, argument count, and schema complexity. \emph{Consistency} compares $K$ low-cost stochastic proposals by exact tool agreement and argument-level normalized agreement. A black-box variant can operate without logits, analogous in spirit to sampling-based uncertainty \citep{manakul2023selfcheck}. \emph{State signals} encode trajectory length, previous tool errors, repeated actions, empty/ambiguous observations, and remaining retry budget. Finally, \emph{risk signals} mark read-only versus state-changing tools, destructive/open-world hints when available, and a deployment-defined severity class. Appendix~\ref{app:risk-taxonomy} gives concrete worked examples for each risk tier. MCP annotations can provide interoperability hints, but because such annotations need not be trustworthy, the implementation must combine them with a local allowlist/policy rather than treating remote metadata as authoritative \citep{mcp2025}.

A lightweight gate $g_\theta(\phi_t)\in[0,1]$ estimates $p_t=\Pr(y_t=1\mid\phi_t)$, where $y_t=1$ denotes that executing the SLM proposal would be judged incorrect by the benchmark/environment. We fit logistic regression, gradient-boosted trees, and a small MLP; the primary model is selected based on validation AUROC and calibration error, not on downstream test performance. Temperature or isotonic calibration is fitted on a disjoint calibration split. This separation matters because a router that ranks failures well but is miscalibrated can violate an escalation budget. This calibration is purely empirical and carries no finite-sample misrouting bound, unlike distribution-free conformal-style calibration (Appendix~\ref{sec:limitations}).

\begin{figure}[t]
\centering
\resizebox{\linewidth}{!}{%
\begin{tikzpicture}[
    font=\scriptsize,
    node distance=3mm and 5mm,
    box/.style={
        draw,
        rounded corners,
        align=center,
        minimum height=5mm,
        inner sep=2pt
    },
    arr/.style={-{Latex[length=2mm]},thick}
]
\node[box] (trace)
    {SLM traces\\$(s_t,a_t,o_{t+1})$};
\node[box,right=of trace] (label)
    {Counterfactual label\\would $a_t$ succeed?};
\node[box,right=of label] (features)
    {Features $\phi_t$\\UQ + schema + state + risk};
\node[box,right=of features] (gate)
    {Calibrated gate\\$g_\theta(\phi_t)=\hat p_{\rm fail}$};
\draw[arr] (trace) -- (label);
\draw[arr] (label) -- (features);
\draw[arr] (features) -- (gate);
\node[box,below=5mm of features] (policy)
    {Budget/risk policy\\
    $L$ if $p<\tau_L$;\;
    $V$ if $\tau_L\le p<\tau_C$;\;
    $C$ otherwise};
\node[box,left=of policy] (constraints)
    {Deployment constraints\\
    cloud budget $B_C$\\
    latency/energy/privacy};
\node[box,right=of policy] (out)
    {Pareto frontier\\
    success vs.\ cloud/latency/energy};
\draw[arr] (gate.south) |- ($(policy.north)+(0,3mm)$) -- (policy.north);
\draw[arr] (constraints) -- (policy);
\draw[arr] (policy) -- (out);
\end{tikzpicture}%
}
\caption{Training and deployment. Failure labels supervise only the gate.
Thresholds are selected on validation data to satisfy a cloud-call/risk
budget, enabling deployment-time trade-offs without retraining either
language model.}
\label{fig:training}
\end{figure}
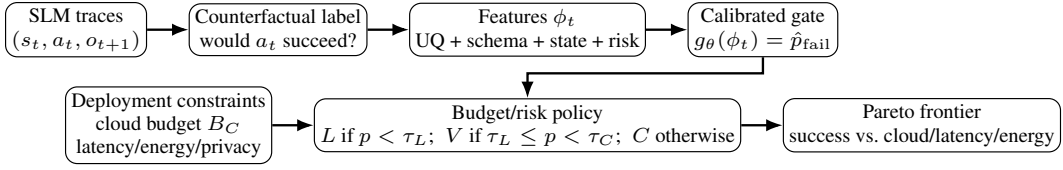

\paragraph{Budget- and risk-aware decision rule.} A binary threshold is insufficient because local verification may be much cheaper than cloud escalation. We minimize expected deployment loss
\begin{equation}
\mathcal{L}(\pi)=\mathbb{E}\!\left[\ell_{task}+\lambda_C C+\lambda_L D+\lambda_E E+\lambda_P P\right],
\label{eq:loss}
\end{equation}
where $C$ is cloud usage/cost, $D$ latency, $E$ local energy, and $P$ privacy exposure (e.g., normalized remote-context tokens). In a budgeted form, maximize task success subject to $\mathbb{E}[C]\le B_C$ and optional latency/energy/privacy constraints. For an interpretable default policy, choose thresholds $\tau_L<\tau_C$: low estimated failure probability executes locally; the middle region requests one bounded local verification/retry; the high region escalates. Risk shifts thresholds monotonically: state-changing/destructive actions require greater confidence for local execution. Thresholds are chosen only on validation data and are empirical rather than backed by a distribution-free coverage guarantee. Figure~\ref{fig:training} shows the training/deployment separation.

\paragraph{Handoff context minimization.} Escalation should not automatically transmit the entire local history. The cloud request includes the user subgoal, the current tool schemas needed for the decision, a compact state summary, the proposed SLM action and uncertainty evidence, and only the tool observations needed to resolve the step. We evaluate full-context and minimized-context handoffs. This turns remote-context exposure into a measurable system variable rather than a qualitative claim.

\paragraph{Algorithm (see Algorithm~\ref{alg:stepgate}).}
For each step, (i) obtain one SLM proposal and cheap intrinsic features, (ii) if the gate is clearly below $\tau_L$, execute, (iii) otherwise acquire optional $K-1$ samples and recompute consistency features, (iv) select local verification or cloud handoff using calibrated risk-adjusted thresholds, and (v) execute exactly one selected action and continue. Adaptive feature acquisition avoids paying for self-consistency on obviously easy steps. A maximum retry count prevents loops. The policy abstains or requests user clarification if neither the local nor the cloud proposal satisfies a deployment safety predicate. Local self-verification here is limited to sample-based consistency plus one bounded revision or retry.

\section{Experimental results}
\label{sec:experiments}

\paragraph{Setup.} We evaluate Qwen2.5-0.5B, 1.5B, and 3B local models against a Qwen2.5-7B 4-bit strong model {, which stands in for the cloud LLM. All results therefore concern one model family and one stronger backend}. The original BFCL-derived function-calling experiment uses 100 tasks with a held-out 52-task test partition. We separate gate training/calibration and threshold selection from that test split, and select the 30\% cloud target before test evaluation. We additionally evaluate the 1.5B/7B configuration on 100 held-out multi-turn trajectories (mean 4.1 agent steps, up to 8) in a sandboxed calendar/email/file/search tool environment with scripted observations, where routing is reconsidered after every step. Appendix~\ref{app:multiturn-setup} gives the full task composition, tool inventory, and exact baseline configurations (including the query-level router) needed to reproduce this split. Our multi-turn evaluation reports trajectory success, action success, cloud-action rate, and successful mixed local/cloud trajectories. Risk tiers remain research annotations. We do not assume the strong model is correct.  {All methods are run on the same tasks, so comparisons are paired. We report paired bootstrap 95\% intervals (10{,}000 resamples over task IDs) and exact McNemar tests on binary trajectory success.}

\paragraph{Baselines and metrics.} We compare local-only, strong-only, and random escalation matched to the realized cloud budget, a query-level router, self-consistency routing, and \method. Task success and cloud escalation are the primary focus. We additionally report wrong risk-tier-$\geq$R2 actions, measured wall-clock latency, remote tokens, and failure-predictor discrimination/calibration. Random routing is averaged over five seeds where reported in the routing sweep. The oracle in supplementary results is diagnostic and non-deployable. We do not compare against risk-controlled routers with formal guarantees (e.g., RACER \citep{hao2026racer}), since adapting its single-turn, fixed-pool formulation into a step-level baseline is nontrivial (Appendix~\ref{sec:limitations}).  {In the single-step test, realized escalation differs across routers (30.8\% \method, 33.8\% random, 42.3\% query-level, 38.5\% self-consistency). Appendix~\ref{app:budget} lists realized rates next to every result and adds rate-matched variants of random, query-level, and self-consistency routing at 30.8\%. In the multi-turn test, all methods except self-consistency (38.0\%) operate at 30.0\% cloud actions.}

\begin{table}[t]
\centering\small
\caption{Main results: qwen2.5-1.5b (local) / qwen2.5-7b 4-bit (strong), real 100-task BFCL subset, 52 test tasks. Cloud rate budget-matched at 30\% target (threshold/model chosen on calibration split only, applied once to test). ``Wrong risky'' = \% of executed test actions that were both incorrect and risk-tier $\geq$R2 (our taxonomy has no R3 action type in this task set).  {Realized cloud rates differ across routers and rate-matched variants are in Appendix~\ref{app:budget}.}}
\label{tab:main}
\setlength{\tabcolsep}{3pt}
\begin{tabular}{lccccc}
\toprule
System & Task succ. $\uparrow$ & Cloud \% $\downarrow$ & Wrong risky $\downarrow$ & Latency $\downarrow$ & Remote tok. $\downarrow$\\
\midrule
SLM only & 67.3 & 0 & 15.4 & 0.702 & 0\\
Strong only & 92.3 & 100 & 3.8 & 2.255 & 475.0\\
Random @ budget & 75.4 & 33.8 & 10.8 & 0.823 & 158.5\\
Query-level router & 73.1 & 42.3 & 15.4 & 1.052 & 266.2\\
Self-consistency & 82.7 & 38.5 & 7.7 & 0.829 & 175.2\\
\textbf{\method} & 82.7 & 30.8 & 3.8 & 0.777 & 134.5\\
\bottomrule
\end{tabular}
\end{table}

\paragraph{Main result.} At the preselected 30\% cloud budget for the 1.5B/7B pair, \method{} reaches 82.7\% task success with 30.8\% escalation (Table~\ref{tab:main}). This represents a +15.4 percentage point improvement over local-only and +7.3 points over random escalation  {(which uses more cloud steps: 33.8\%)}, while reducing cloud escalations by 69.2\% compared to strong-only. While strong-only reaches 92.3\% task success at 100\% remote execution, \method{} closes over 60\% of this headroom while running nearly 70\% of actions locally. Compared with the query-level router, \method{} achieves +9.6 points higher task success while escalating fewer actions (30.8\% versus 42.3\%) {. This baseline also escalates more, so the comparison favors it rather than \method}. Self-consistency matches \method's 82.7\% success rate but requires 38.5\% escalation  {and has a higher wrong-risky rate (7.7\% vs.\ 3.8\%)}.  {On this 52-task split one task equals 1.9 points, so the +7.3 and +9.6 gaps correspond to about four and five tasks, and the individual 95\% Wilson intervals overlap (Appendix~\ref{app:budget}). We read the single-step result as supporting evidence and rest our main claim on the multi-turn comparison.} \method{} executed a risky wrong action in only 3.8\% of cases, compared to 15.4\% for local-only and 10.8\% for random escalation  {(2, 8, and 6 out of 52 steps). While the small sample size limits strong conclusions, this suggests uncertainty gating helps prevent high-consequence mistakes on this benchmark}.

\begin{table}[t]
\centering\scriptsize
\caption{Multi-turn results for Qwen2.5-1.5B local and Qwen2.5-7B strong models, evaluated on the 100-task multi-turn split described in Appendix~\ref{app:multiturn-setup}. ``Mixed success'' is the share of trajectories that combined at least one local and one cloud action and still succeeded.  {$^\ddagger$Unpaired 95\% Wilson interval for each method's trajectory success ($n{=}100$; random escalation is the five-seed mean, so its interval is approximate). Paired differences are given in the text.}}
\label{tab:multiturn}
\setlength{\tabcolsep}{2.4pt}
\begin{tabular}{lrrrrr}
\toprule
Method & Traj. succ. $\uparrow$ &  {95\% CI$^\ddagger$} & Action succ. $\uparrow$ & Cloud actions $\downarrow$ & Mixed success $\uparrow$\\
\midrule
Local-only & 48.0\% &  {[38.5, 57.7]} & 70.5\% & 0.0\% & --\\
Random escalation & 60.0\% &  {[50.2, 69.1]} & 78.2\% & 30.0\% & 34.0\%\\
Query-level router & 57.0\% &  {[47.2, 66.3]} & 77.1\% & 30.0\% & 0.0\%\\
Self-consistency & 66.0\% &  {[56.3, 74.5]} & 82.6\% & 38.0\% & 48.0\%\\
\textbf{\method} & \textbf{69.0\%} &  {[59.4, 77.2]} & \textbf{84.0\%} & 30.0\% & \textbf{55.0\%}\\
Strong-only & 82.0\% &  {[73.3, 88.3]} & 91.0\% & 100.0\% & --\\
\bottomrule
\end{tabular}
\end{table}

\paragraph{Multi-turn handoff.} Table~\ref{tab:multiturn} directly tests the step-level motivation. At 30.0\% cloud actions, \method{} reaches 69.0\% trajectory success, +21.0 points over local-only, +9.0 over random escalation at the same cloud-action rate, and +12.0 over query-level routing. Action success reaches 84.0\%, with 55.0\% of tasks successfully completed through collaborative local/cloud execution. By contrast, self-consistency attains 66.0\% trajectory success but consumes 38.0\% cloud actions. While a pure cloud strong model attains 82.0\% trajectory success, \method{} closes over 60\% of this headroom while preserving local autonomy for 70\% of steps. 

 {\textbf{Paired analysis.} Over the 100 identical tasks, the trajectory-success difference of \method{} (paired bootstrap 95\% CI, exact two-sided McNemar $p$) is $+21.0$ {[11.0, 31.0]}, $p{=}${0.0003} versus local-only. Compared to random escalation, the difference is $+9.0$ {[$-$0.6, 18.8]}, $p{=}${0.108} (paired against each of the five seeds, with the mean interval reported). It achieves $+12.0$ {[2.0, 22.0]}, $p{=}${0.029} versus query-level routing, and $+3.0$ {[$-$5.0, 11.0]}, $p{=}${0.629} versus self-consistency, which uses 8 percentage points more cloud actions. Consequently, the gains over local-only and query-level routing exclude zero, whereas those over random escalation and self-consistency do not, so we treat the latter two comparisons as suggestive. The unpaired intervals in Table~\ref{tab:multiturn} overlap across the four routed methods, which is consistent with this interpretation.}

\begin{figure}[h]
\centering
\includegraphics[width=\linewidth]{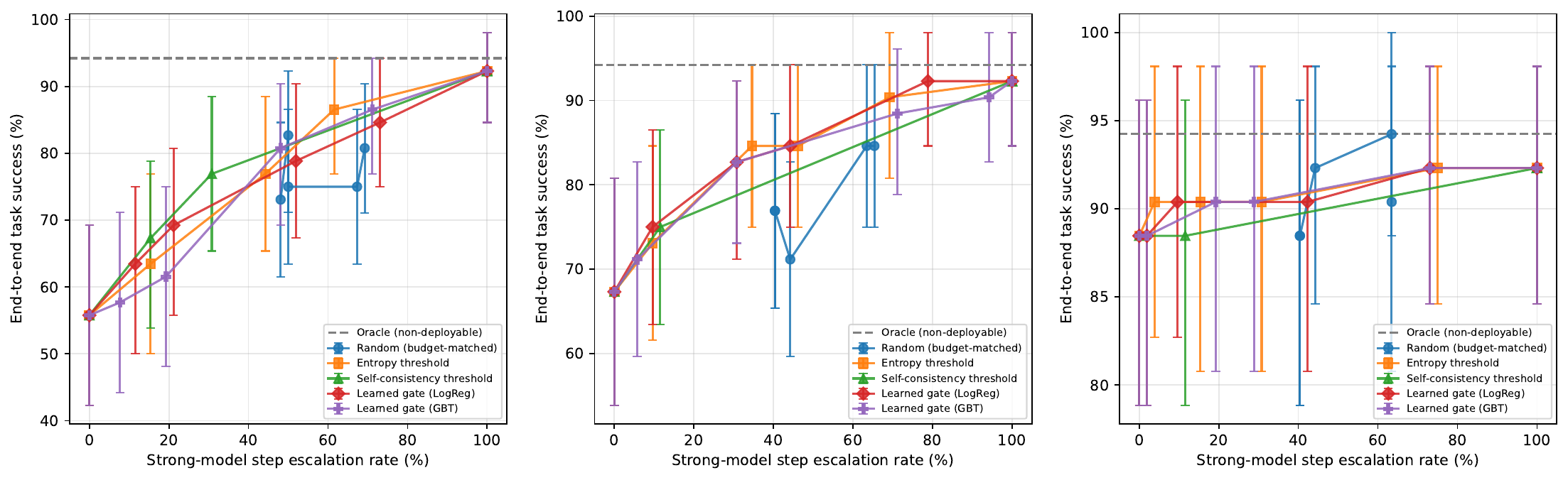}
\caption{Measured end-to-end task success versus realized strong-model step escalation rate for Qwen2.5-0.5B, 1.5B, and 3B local models. Error bars are the task-level 95\% confidence intervals from the executed routing sweep. The dashed oracle is non-deployable. The plot exposes an important scale effect: routing offers substantially more headroom for weaker local models, whereas the 3B local baseline is already close to the strong-model ceiling.}
\label{fig:quality_escalation}
\end{figure}

\paragraph{Model scale and routing curves.} Figure~\ref{fig:quality_escalation}  summarizes the measured success-escalation trade-off across all three local-model sizes. Local-only success rises from 55.8\% (0.5B) to 67.3\% (1.5B) and 88.5\% (3B), while strong-only is 92.3\%. At the 50\%-target sweep, learned routing provides substantial headroom for 1.5B (84.6\% logistic; 88.5\% gradient-boosted trees), but yields diminishing returns for 3B (90.4\% vs. an 88.5\% local baseline and a 90.8\% random escalation baseline). Consequently, learned routing offers the greatest leverage when the local model leaves a substantial recoverable-error margin. As local SLM capacity approaches the strong model's ceiling, the value of selective handoff naturally saturates.

\paragraph{Calibration and ablations.} For the 1.5B full feature set, logistic regression obtains AUROC 0.829, AUPRC 0.832, Brier 0.144, and ECE 0.212. Gradient-boosted trees achieve an AUROC of 0.854 and an ECE of 0.121. The ablation at the 30\% target is not monotonic: removing risk metadata improves logistic AUROC to 0.886 and downstream task success to 86.5\% at 34.6\% realized escalation, while removing consistency or schema features leaves task success at 82.7\%. Token uncertainty alone reaches only 76.9\% success at 28.8\% escalation.  {The risk ablation is not budget-matched: the no-risk gate escalates two more of 52 steps (34.6\% vs.\ 30.8\%) and completes two more tasks (86.5\% vs.\ 82.7\%), so the gain is consistent with the extra budget alone and neither supports nor refutes a benefit of risk metadata. At an equal 30.8\% budget (threshold re-tuned on validation data) the no-risk gate reaches {82.7\%} task success and a {5.8\%} wrong-R2+ rate, versus 82.7\% and 3.8\% for the full gate (Appendix~\ref{app:budget} and Table~\ref{tab:ablation}): task success is identical, and the wrong-R2+ difference is one step of 52, too small to be conclusive. Because risk tiers are meant to shift thresholds for consequential actions rather than raise average accuracy, we do not claim that risk features improve task success.} We interpret the remaining non-monotonicity as feature interactions and possible overfitting on a 52-task split (Appendix~\ref{sec:limitations}) rather than a clean per-family ranking, arguing for calibration and feature selection rather than assuming that every uncertainty signal helps.

\paragraph{Efficiency.} Measured median local latency increases with SLM size: 0.301s (0.5B), 0.702s (1.5B), and 1.152s (3B), with corresponding local peak memory of 990, 3003, and 6058~MiB. The measured strong-model median latency is 1.528s. For the main 1.5B operating point, \method's measured end-to-end latency is 0.777s versus 2.255s for strong-only, and remote tokens are 134.5 versus 475.0 (a disclosure-volume proxy, not a privacy guarantee). Local energy measurements are hardware-specific and are reported only in the appendix. They should not be generalized to other accelerators.

\paragraph{Reproducibility and statistical scope.} The routing sweep reports task-level 95\% bootstrap confidence intervals. For instance, the 1.5B local-only estimate is 67.3\% [53.8, 80.8] versus 92.3\% [84.6, 98.1] for strong-only.  {Single-step differences between routers are evaluated on 52 tasks and interpreted descriptively (Appendix~\ref{app:budget}). In contrast, multi-turn comparisons follow the paired procedure described above. Neither setting supports generalized claims beyond the evaluated benchmarks.} The empirical findings reflect the evaluation of in-distribution benchmark tasks, scripted multi-turn observations, and the specified open-weight model pairings.

\section{Conclusion}

We presented \method, an uncertainty-aware, step-level handoff mechanism for SLM agents. On the 52-task single-step benchmark, the 1.5B/7B pair improves task success from 67.3\% (local-only) to 82.7\% with only 30.8\% escalation. In multi-turn evaluations, \method{} attains 69.0\% trajectory success and 84.0\% action success using a 30.0\% cloud budget, outperforming local-only (48.0\%/70.5\%) and random escalation (60.0\%/78.2\%), with 55.0\% of trajectories successfully completed via mixed local/cloud execution. In comparison, strong-only reaches 82.0\% trajectory success at the expense of 100\% remote execution.  {These results suggest that selective, step-level handoff offers an effective efficiency--reliability trade-off in this setting. However, they do not establish generality beyond one model family, one stronger backend, and scripted tasks. Furthermore, the risk tiers are empirical annotations rather than formal safety guarantees, remote-token counts do not ensure privacy, and the individual contribution of the risk features remains unestablished.} Extending our empirical framework with finite-sample, distribution-free risk guarantees represents a compelling direction for future research.

\appendix
\section{Technical Appendix}

 {\subsection{Budget-matched single-step comparison and equal-budget risk ablation}
\label{app:budget}
Table~\ref{tab:budget} lists, for the 1.5B/7B pair on the 52-task single-step test split, the realized escalation rate next to each result, with unpaired 95\% Wilson intervals computed from the reported counts (one task $=1.9$ points). Rows marked ``matched'' re-tune the router's threshold (or, for random, its escalation probability) on validation data so that the realized test escalation equals \method's 30.8\% (16 of 52 steps). We retune the no-risk gate the same way. ``As reported'' rows are the measured values of Table~\ref{tab:main} and the original ablation. Success--escalation curves for each router over the 10--50\% range should be read together with this table. Wrong-R2+ is the executed wrong risky-action rate on actions annotated R2 or above (a benchmark annotation, not a real-world harm rate).}

\begin{table}[h]
\centering\scriptsize
\caption{ {Single-step test (52 tasks), Qwen2.5-1.5B/7B. $^\dagger$Five-seed mean; interval approximate. ``As reported'' rows use the thresholds of the original experiment. ``matched'' rows are re-tuned on validation data to 30.8\% realized escalation.}}
\label{tab:budget}
\setlength{\tabcolsep}{3pt}
\begin{tabular}{lcccc}
\toprule
 {Method} &  {Realized esc.} &  {Task success} &  {95\% Wilson CI} &  {Wrong R2+}\\
\midrule
 {Local-only} &  {0.0\%} &  {67.3\%} &  {[53.8, 78.5]} &  {15.4\%}\\
 {Random escalation, as reported$^\dagger$} &  {33.8\%} &  {75.4\%} &  {[62.2, 85.1]} &  {10.8\%}\\
 {Random escalation, matched$^\dagger$} &  {30.8\%} & {74.7\%} & {[61.5, 84.5]} & {11.2\%}\\
 {\method{} (full)} &  {30.8\%} &  {82.7\%} &  {[70.3, 90.6]} &  {3.8\%}\\
 {\method{} no-risk, as reported} &  {34.6\%} &  {86.5\%} &  {[74.7, 93.3]} & {3.8\%}\\
 {\method{} no-risk, matched} &  {30.8\%} & {82.7\%} & {[70.3, 90.6]} & {5.8\%}\\
 {Query-level, as reported} &  {42.3\%} &  {73.1\%} &  {[59.7, 83.2]} &  {15.4\%}\\
 {Query-level, matched} &  {30.8\%} & {69.2\%} & {[55.7, 80.1]} & {15.4\%}\\
 {Self-consistency, as reported} &  {38.5\%} &  {82.7\%} &  {[70.3, 90.6]} &  {7.7\%}\\
 {Self-consistency, matched} &  {30.8\%} & {78.8\%} & {[66.0, 87.8]} & {9.6\%}\\
 {Strong-only} &  {100.0\%} &  {92.3\%} &  {[81.8, 97.0]} &  {3.8\%}\\
\bottomrule
\end{tabular}
\end{table}

This appendix records implementation details and supporting measurements omitted from the main text.

\paragraph{Single-step feature ablation.} The detailed ablation table is moved here to prioritize the multi-turn trajectory result in the main paper.
\begin{table}[b]
\centering\small
\caption{Ablation of the calibrated failure predictor (logistic regression), qwen2.5-1.5b, real BFCL subset (n=52 test tasks). Cloud budget fixed at 30\% (threshold chosen on calibration split only).  {The risk gate escalates 2 more of 52 steps than the full gate. The last-but-one row re-tunes its threshold to the same 30.8\% budget (AUROC/ECE are unchanged because the scoring model is the same).}}
\label{tab:ablation}
\setlength{\tabcolsep}{4pt}
\begin{tabular}{lcccc}
\toprule
Gate features & AUROC $\uparrow$ & ECE $\downarrow$ & Task succ. $\uparrow$ & Cloud \%\\
\midrule
All (\method) & 0.829 & 0.212 & 82.7 & 30.8\\
$-$ consistency & 0.832 & 0.231 & 82.7 & 28.8\\
$-$ schema/arguments & 0.829 & 0.241 & 82.7 & 30.8\\
$-$ risk & 0.886 & 0.119 & 86.5 & 34.6\\
 {$-$ risk, matched budget} &  {0.886} &  {0.119} & {82.7} &  {30.8}\\
Token uncertainty only & 0.565 & 0.078 & 76.9 & 28.8\\
\bottomrule
\end{tabular}
\end{table}

\subsection{Formal decision problem and routing rule}
At agent step $t$, let $s_t = (x, h_t, \mathcal{T}_t)$ denote the agent state containing the user goal $x$, compact trajectory history $h_t$, and candidate tool schemas $\mathcal{T}_t$. The local SLM samples a proposed structured action $a_t \sim p_S(\cdot \mid s_t)$. Let $Y_t \in \{0, 1\}$ denote whether executing $a_t$ would fail under the environment evaluator. A lightweight gate $g_\theta$ estimates failure probability $q_t = P(Y_t = 1 \mid \phi_t)$ from action-aware features $\phi_t = \phi(s_t, a_t)$. Based on $q_t$, action risk tier $r_t$, and remaining resource budget $b_t$, the routing policy $\pi(z_t \mid q_t, r_t, b_t)$ selects an action $z_t \in \{L, V, C\}$ corresponding to local execution ($L$), bounded local verification ($V$), or cloud escalation ($C$). The optimal deployment policy solves the multi-objective loss minimization:
\begin{equation}
\min_\pi \; \mathbb{E}_{\pi}\!\left[\ell_{\mathrm{task}} + \lambda_C C + \lambda_D D + \lambda_E E + \lambda_P P\right],
\label{eq:routing_objective}
\end{equation}
where $C$ denotes cloud invocations, $D$ latency, $E$ local energy, and $P$ remote context exposure. For deployment, we instantiate an interpretable thresholding policy defined by risk- and budget-conditioned cutoffs $\tau_L(r_t, b_t) < \tau_C(r_t, b_t)$:
\begin{equation}
z_t = 
\begin{cases}
L & \text{if } q_t \le \tau_L(r_t, b_t), \\
V & \text{if } \tau_L(r_t, b_t) < q_t < \tau_C(r_t, b_t), \\
C & \text{if } q_t \ge \tau_C(r_t, b_t).
\end{cases}
\end{equation}
Thresholds are calibrated exclusively on validation trajectories. For state-changing or high-risk actions ($r_t \ge \text{R2}$), thresholds shift conservatively toward verification and escalation, while deterministic external authorization policies retain the ability to override routing decisions.

\paragraph{Adaptive feature acquisition.} Cheap features are computed from the first local decode. Expensive features such as $K$-sample consistency are acquired only when the cheap gate score lies in an ambiguity band $[\tau_A^-,\tau_A^+]$. This prevents uncertainty estimation from consuming more compute than it saves. All thresholds, $K$, and ambiguity bands are selected without test labels.

\begin{figure}[h]
\centering
\resizebox{\linewidth}{!}{%
\begin{tikzpicture}[font=\scriptsize,node distance=4mm and 7mm,
box/.style={draw,rounded corners,align=center,minimum height=6mm,inner sep=3pt},
arr/.style={-{Latex[length=2mm]},thick}]
\node[box] (p) {SLM proposal\\$a_t$};
\node[box,right=of p] (cheap) {Cheap features\\entropy, margin, schema};
\node[box,right=of cheap] (g1) {Stage-1 gate\\$q_t^{(1)}$};
\node[box,above right=5mm and 8mm of g1] (exec) {local execute};
\node[box,right=8mm of g1] (extra) {Acquire $K$ samples\\+ state consistency};
\node[box,right=of extra] (g2) {Calibrated gate\\$q_t^{(2)}$};
\node[box,above right=5mm and 7mm of g2] (ver) {local verify};
\node[box,below right=5mm and 7mm of g2] (cl) {cloud handoff};
\draw[arr] (p)--(cheap); \draw[arr] (cheap)--(g1); \draw[arr] (g1)--node[above,sloped]{confident}(exec);
\draw[arr] (g1)--node[above]{ambiguous}(extra); \draw[arr] (extra)--(g2); \draw[arr] (g2)--(ver); \draw[arr] (g2)--(cl);
\end{tikzpicture}%
}
\caption{Adaptive uncertainty acquisition. Expensive self-consistency is computed only for ambiguous actions. This appendix figure makes explicit which uncertainty signals incur additional local inference. Note that this diagram covers only sample-based consistency and a single bounded revision.}
\label{fig:adaptive_appendix}
\end{figure}
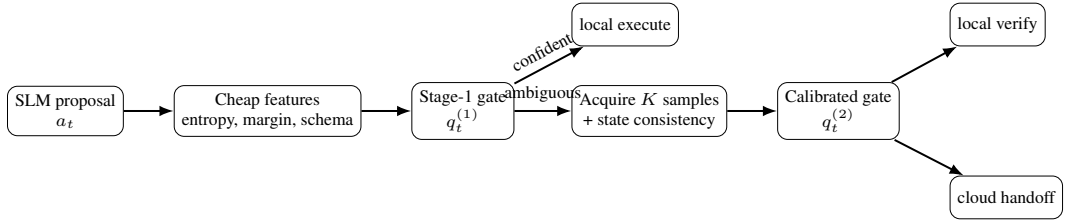

\paragraph{Reference pseudocode.} Algorithm~\ref{alg:stepgate} gives the complete per-step decision procedure summarized in Section~3 (``Algorithm''), spelling out the exact branching between local execution, the ambiguity-band retry/verification path, and cloud escalation, together with the safety rule that governs authorization.
\begin{algorithm}[t]
\caption{\method{} inference at one agent step}
\label{alg:stepgate}
\begin{algorithmic}[1]
\Require state $s_t$, local model $S$, strong model $M$, gate $g_\theta$, thresholds $\tau_L, \tau_C$, maximum retry $R$
\Ensure executed action, updated state $s_{t+1}$
\State Decode one structured proposal $a_t$ from $S$ and compute cheap features $\phi_t^{c}$
\State Compute $q_t^{(1)} = g_\theta(\phi_t^{c})$
\If{risk-adjusted $q_t^{(1)} \le \tau_L$}
    \State Execute $a_t$
\ElsIf{$q_t^{(1)}$ lies in the ambiguity band}
    \State Sample up to $K-1$ additional local proposals
    \State Compute tool/argument agreement and update $q_t$
    \If{$q_t < \tau_C$ \textbf{and} retry budget remains}
        \State Perform one bounded local verification/revision
        \State Re-score the revised action
    \Else
        \State Construct a minimized handoff context and query $M$
        \State Validate the returned structured action against the schema and deployment policy
    \EndIf
\Else
    \State Construct a minimized handoff context and query $M$
    \State Validate the returned structured action against the schema and deployment policy
\EndIf
\State Execute at most one authorized action, record the observation and routing telemetry, and advance to $s_{t+1}$
\Statex \textbf{Safety rule:} schema validity or low uncertainty never grants authorization; application policy may force abstention or human confirmation.
\end{algorithmic}
\end{algorithm}
\subsection{Feature definitions, calibration, and labels}
The frozen implementation should expose each feature separately so that ablations can be reproduced. \emph{Intrinsic} features include mean token negative log-likelihood over the tool name and arguments, maximum token entropy, sequence length, and the probability margin between the top two tool names when tokenization permits; this family is conceptually related to token-level epistemic uncertainty estimators such as TokUR \citep{zhang2026tokur}, though we use plain decode-time statistics rather than weight-perturbation ensembles. \emph{Structural} features include JSON/schema validity, number and type of required arguments, enum constraints, nesting depth, and whether argument values are copied from or grounded in the observed state. \emph{Consistency} features compare $K$ stochastic proposals using exact tool agreement, normalized argument agreement, and execution-equivalence where a deterministic sandbox evaluator is available. \emph{State} features include trajectory depth, number of prior failures, whether the current action depends on a previous observation, and candidate-tool count. \emph{Risk} features are externally supplied and never inferred solely from the model.

For each local proposal, the binary failure label is computed by the benchmark/environment before any cloud correction. Multi-turn tasks additionally retain trajectory-level success. The gate is trained only on local-model proposals from a calibration/training partition, preventing labels from strong-model outputs from leaking into the failure predictor. We recommend logistic regression and gradient-boosted trees as transparent baselines before any neural gate. Calibration should compare temperature scaling, isotonic regression, and uncalibrated scores using ECE and Brier score. We freeze the selected calibrator on validation data. We note that this calibration protocol targets good ranking and low average ECE, not a guaranteed per-deployment bound on misrouting risk. Readers wanting the latter should consult finite-sample conformal risk-control procedures such as RACER's $\alpha$-VOR calibration \citep{hao2026racer} or classical conformal prediction \citep{vovk2005conformal}, which we discuss as an open extension in Appendix~\ref{sec:limitations}.

\begin{table}[h]
\centering\small
\caption{Feature taxonomy and reporting overview. ``Extra decode'' identifies signals that incur additional local generation and therefore must be accounted for in latency/energy.}
\label{tab:features_appendix}
\renewcommand{\arraystretch}{1.15}
\begin{tabular}{p{0.24\linewidth}p{0.50\linewidth}c}
\toprule
\textbf{Family} & \textbf{Representative signals} & \textbf{Extra decode?} \\
\midrule
Intrinsic     & token NLL, entropy, tool-choice margin, output length & No \\
Structural    & schema validity, argument count/type, enum/nesting complexity & No \\
Consistency   & tool agreement, argument agreement, execution equivalence & Yes \\
State/history$^\dagger$ & trajectory depth, previous errors, observation dependence & No \\
Risk          & read-only/state-changing/destructive/external side effect & No \\
\bottomrule
\multicolumn{3}{l}{\footnotesize $^\dagger$Evaluated exclusively within multi-turn sequential trajectories.}
\end{tabular}
\end{table}

\subsection{Background: conformal prediction and risk control}
\label{app:conformal-background}
For readers unfamiliar with the guarantee referenced in Appendix~\ref{sec:limitations} and Appendix~\ref{app:extended-related}: conformal prediction constructs prediction sets that provably contain the true label with a user-specified probability, using only exchangeability of calibration and test data and no distributional assumptions on the underlying model \citep{vovk2005conformal}. Conformal risk control generalizes this from set coverage to bounding the expected value of any monotone loss function by calibrating a single threshold on a held-out calibration set. RACER's $\alpha$-VOR calibration (Appendix~\ref{app:extended-related}) is an instance of this idea specialized to model routing: the ``loss'' is whether the selected model set excludes every acceptable model, and the guaranteed bound holds for any finite calibration sample size, not just asymptotically. In contrast, \method's gate calibration (temperature/isotonic scaling; Appendix~B) targets good \emph{average} calibration (low ECE/Brier) on a validation split, which is a different and weaker property: it does not bound the probability of misrouting on any particular deployment sample, and the current implementation does not verify or rely on an exchangeability assumption across trajectory steps, which conformal-style guarantees would require.

\subsection{Risk taxonomy: concrete examples}
\label{app:risk-taxonomy}
Section~3 and Appendix~\ref{sec:limitations} describe the risk tiers R0--R3 as research annotations rather than a security standard. Table~\ref{tab:risk-examples} gives representative tool calls for each tier, drawn from the calendar/email/file/search tool families used in our multi-turn environment (Appendix~\ref{app:multiturn-setup}), to make the annotation rule concrete for readers who wish to reproduce or extend it.

\begin{table}[h]
\centering\small
\caption{Worked examples of the risk taxonomy used to annotate tool calls. Tier assignment follows reversibility and blast radius, not tool name alone: two calls to the same underlying tool can receive different tiers depending on arguments (e.g., a narrowly scoped read vs.\ a bulk export).}
\label{tab:risk-examples}
\begin{tabular}{p{0.08\linewidth}p{0.24\linewidth}p{0.6\linewidth}}
\toprule
Tier & Description & Example calls\\
\midrule
R0 & Read-only, no external mutation & \texttt{search\_events(range=...)}; \texttt{get\_email(id=...)}; \texttt{list\_files(folder=...)}; \texttt{web\_search(query=...)}\\
R1 & Reversible, low-impact mutation & \texttt{create\_draft\_email(...)} (not sent); \texttt{add\_calendar\_event(private=True)}; \texttt{rename\_file(...)} within the user's own workspace\\
R2 & Externally visible or financially/privacy-relevant mutation & \texttt{send\_email(to=external\_recipient, ...)}; \texttt{share\_file(recipient=external, permission=view)}; \texttt{reschedule\_meeting(notify\_attendees=True)}\\
R3 & Destructive or difficult-to-reverse action & \texttt{delete\_event(recurring=True)}; \texttt{permanently\_delete\_file(...)}; \texttt{bulk\_delete\_emails(filter=...)}; \texttt{cancel\_meeting(notify\_all=True)} for an externally organized event\\
\bottomrule
\end{tabular}
\end{table}

Two practical notes follow from these examples. First, tier assignment is a function of the \emph{call} (tool plus bound arguments), not just the tool name, since the same endpoint can be R0 in a narrowly scoped read and R2--R3 when arguments broaden its blast radius (e.g., a filtered vs.\ unfiltered bulk delete). Second, we fixed tier boundaries in our environment before test evaluation using the same deterministic rule set described in Section~4 of the main text and did not tune them against downstream task success to avoid the annotation itself becoming a source of test-set leakage.

\subsection{Multi-turn environment and baseline specification}
\label{app:multiturn-setup}
This appendix specifies the multi-turn evaluation referenced in Section~4 (Setup) to support replication.

\paragraph{Environment and task composition.} The multi-turn split contains 100 held-out tasks executed in a sandboxed, scripted tool-use environment (not a live API), spanning four tool families: calendar (7 tools, e.g., search/create/update/delete events), email (6 tools, e.g., search/read/draft/send), file storage (5 tools, e.g., list/read/rename/share/delete), and web/knowledge search (2 tools). Each task specifies a natural-language user goal, an initial environment state (populated calendar/inbox/file tree), and a deterministic evaluator that checks final environment state and/or the sequence of executed actions against task-specific success criteria. Trajectories average 4.1 agent steps (min 1, max 8); a step ends in an executed tool call, and the trajectory ends when the evaluator's goal condition is met, the agent abstains, or the step budget (8) is exhausted. Task difficulty varies along three axes we tag but do not stratify results by in the main text: whether disambiguation is required after an observation (e.g., multiple matching events), whether at least one R2/R3 action is required to complete the goal, and whether an early tool result invalidates the agent's initial plan (forcing re-planning). Tasks were authored to cover all four tool families and all four risk tiers (Appendix~\ref{app:risk-taxonomy}) at least twice each; no task or its paraphrase appears in the gate's training or calibration data.

\paragraph{Baseline configuration.} All compared methods share the same underlying Qwen2.5-1.5B local and Qwen2.5-7B (4-bit) strong models, the same tool schemas, and the same evaluator; they differ only in the routing decision at each step. \emph{Local-only} always executes the 1.5B model's proposal. \emph{Strong-only} always executes the 7B model's proposal. \emph{Random escalation} draws the routing decision i.i.d.\ per step from a Bernoulli distribution whose parameter is tuned so the realized cloud-action rate matches \method's 30.0\% (results averaged over five seeds). \emph{Query-level router} computes the gate's features once from the initial user goal and tool schemas (i.e., at $t=0$, before any observation) and commits to local or cloud execution for every step of the trajectory; it cannot react to later observations, which is the central contrast with \method. \emph{Self-consistency routing} re-samples $K=5$ local proposals at every step and escalates when tool-choice agreement across samples falls below a fixed threshold tuned on the calibration split. \emph{\method} is the full step-level, risk-adjusted gate described in Section~3, re-evaluated after every observation. All methods use the same minimized handoff-context construction (Section~3, ``Handoff context minimization'') when escalating, so the latency/token differences in Section~4 are attributable to the escalation policy rather than to context size.

\paragraph{Statistical scope.}  {With 100 tasks, a 21-point gap corresponds to 21 additional successful trajectories out of 100. We report paired outcomes on identical task IDs: a paired bootstrap (10{,}000 resamples over tasks) for the difference in trajectory success, and an exact two-sided McNemar test on the discordant pairs. Random escalation is paired against each of its five seeds, and the mean interval and per-seed $p$-values are reported. Per-task outcomes and the analysis script are released with the artifact. The unpaired Wilson intervals in Table~\ref{tab:multiturn} are shown for reference only and are not a substitute for the paired test. A larger, pre-registered multi-turn benchmark and stratification by the three difficulty tags above remain future work (Appendix~\ref{sec:limitations}).}

\subsection{Dataset construction, splits, and leakage prevention}
The primary benchmark should be pinned to an immutable BFCL release or commit and evaluated using its official evaluator wherever possible. Report every included category separately and state all exclusions. If an execution-safe risk subset is created, annotation rules must be deterministic and released. Appendix~\ref{app:risk-taxonomy} gives the concrete tier examples used in this work. A suggested taxonomy is R0: read-only/no external mutation. R1 is a reversible low-impact mutation. R2 is an externally visible or financially/privacy-relevant mutation, and R3 is a destructive or difficult-to-reverse action. The taxonomy is a research annotation, not a security standard.

Calibration, threshold selection, and final evaluation must be separated by task IDs and, where feasible, by tool families to test transfer to unseen tools. No final-test outcome may influence feature engineering, calibrator choice, thresholds, prompt selection, or retry count. For distribution-shift experiments, construct fixed transformations before evaluation: paraphrased tool descriptions, increased distractor-tool count, reordered schemas, longer trajectories, and held-out tool families. Preserve exact transformation seeds and generated artifacts. We flag distribution shift and cross-environment transfer (e.g., to ToolSandbox or WebArena-style tasks) as unevaluated in this work.

\subsection{Prompts, handoff packet, and tool safety}
The artifact should contain verbatim system prompts and chat templates. The local actor prompt should request exactly one next action or explicit abstention, not hidden free-form chain-of-thought. The verifier prompt, if used, should receive the user subgoal, proposed action, relevant schema, and the minimal observation needed to judge correctness. Evaluation should score observable actions rather than private reasoning traces.

The minimized cloud packet should be constructed from explicit fields: (1) current user subgoal; (2) relevant tool names and schemas; (3) compact state summary; (4) proposed local action; (5) uncertainty/risk metadata; and (6) only observations required to resolve the decision. Report both raw remote token count and a field-level disclosure inventory. Sensitive values should be redacted or tokenized whenever the strong model does not need their literal content. Full-context handoff is retained only as an upper-bound baseline for accuracy.

All state-changing tools used in experiments should execute in mocks, sandboxes, or disposable accounts. The gate must not be treated as an access-control system. Before tool execution, an independent deterministic layer should enforce schema validation, allowlists, rate limits, and any required human confirmation. When used, MCP tool annotations are treated as untrusted hints rather than authorization.

\subsection{Baselines and budget matching}
Every routing method must be evaluated at multiple operating points. For a fair comparison, match the number of strong-model calls (or strong-model tokens, in a secondary analysis) by choosing each baseline threshold on validation data. We repeat random escalation across multiple seeds. Query-level routing makes a single decision before the first tool action and retains it throughout the trajectory (Appendix~\ref{app:multiturn-setup} gives the exact procedure used in our multi-turn experiment). Step-level baselines may be re-evaluated after each observation. The oracle sees local failure labels and is reported only as an unattainable upper bound.

Recommended operating points are approximately 0, 10, 25, 50, 75, and 100\% cloud-step usage where attainable. Curves should report all attainable points rather than a single favorable threshold. If API rate limits or costs prevent dense sweeps, cache local proposals first and pre-register the selected operating points before strong-model evaluation.

\subsection{Metrics, uncertainty intervals, and statistical analysis}
Task success is the primary effectiveness metric. Cloud-step rate is the primary resource metric. Secondary metrics include per-action correctness, failure-to-abstain rate, risky wrong-action rate, strong/local token counts, end-to-end latency, peak memory, energy when measurable, and remote-context tokens. Measure router discrimination with AUROC and AUPRC, and calibration with Brier score and ECE. Selective behavior is measured with risk-coverage and success-cloud curves.

For task-level proportions, report paired bootstrap 95\% confidence intervals over task IDs with at least 10,000 resamples in the final artifact. When comparing two routers on the same binary task outcomes, use McNemar's test or a paired randomization test and report the test definition. For learned gates, repeat training/calibration over at least three seeds and distinguish seed variability from task-resampling uncertainty. Do not interpret overlapping/non-overlapping confidence intervals as a substitute for a paired test. Report absolute effects alongside relative improvements. Readers seeking a router with a provable, rather than bootstrap-estimated, bound on misrouting probability should consider adapting finite-sample risk-control procedures such as RACER \citep{hao2026racer} or conformal risk control \citep{vovk2005conformal} to the step-level, stateful setting.

\paragraph{Pareto reporting.} A router is practically useful only if it improves the frontier. For each method, plot task success against cloud-step rate and separately risky wrong-action rate against cloud-step rate. Report the area under the success-cloud curve only as a summary. The full curve remains the primary focus. A budget-matched claim is supported only when the compared methods use the same cloud budget within a declared tolerance.

\subsection{Efficiency and hardware measurement protocol}
Hardware experiments should freeze the device, power mode, OS, drivers, inference runtime, model revision, quantization, context length, batch size, and decoding configuration. Perform a declared warm-up, then measure complete agent steps rather than model kernels alone. Report median and tail latency in addition to mean latency. If energy counters are available, state their source, sampling interval, integration procedure, idle-power treatment, and whether tool execution is included. API latency should be measured separately from local latency because network conditions are not reproducible across sites.

To avoid misleading efficiency claims, account for the additional local samples used by self-consistency and any verifier calls. The principal resource table should therefore include locally generated tokens, strong-model generated/input tokens, total model invocations, and wall-clock time. We report monetary cost only with the exact price schedule and date used for calculation.

\subsection{Additional measured results}
\color{black}
Table~\ref{tab:appendix_models} reports the measured 30\%-budget results across the three local-model sizes. These measurements characterize the evaluated in-distribution setting; robustness to unseen tool families, schema paraphrases, distractor growth, and longer trajectories is outside this study's empirical scope.

\begin{table}[h]
\centering\small
\caption{Per-model results at 30\% fixed cloud budget, real BFCL subset. Three local/strong pairs were evaluated (a single strong model and three SLM sizes).}
\label{tab:appendix_models}
\begin{tabular}{lccccc}
\toprule
Local / strong pair & Params & Task succ. & Cloud \% & Risky errors & Latency\\
\midrule
qwen2.5-0.5b / qwen2.5-7b & 0.5B & 73.1 & 26.9 & 3.8 & 0.351\\
qwen2.5-1.5b / qwen2.5-7b & 1.5B & 82.7 & 30.8 & 3.8 & 0.777\\
qwen2.5-3b / qwen2.5-7b & 3B & 90.4 & 26.9 & 3.8 & 1.153\\
\bottomrule
\end{tabular}
\end{table}

\begin{figure}[h]
\centering
\includegraphics[width=.6\linewidth]{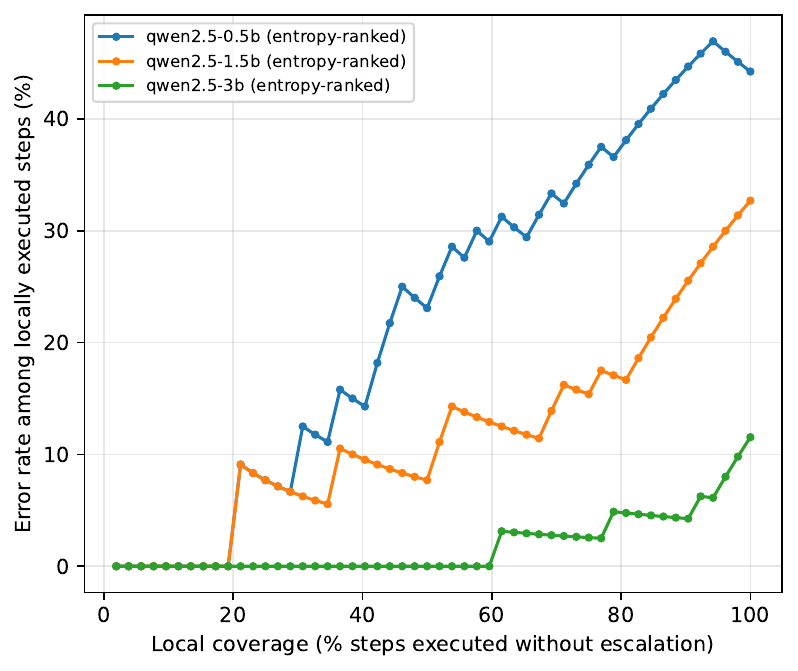}
\caption{Selective risk under entropy-ranked local coverage. The error rate among locally executed steps rises with coverage, most sharply for the 0.5B model and least for the 3B model. This plot evaluates ranking behavior; it is not the learned-gate routing curve.}
\label{fig:selective_risk}
\end{figure}

\begin{figure}[h]
\centering
\includegraphics[width=\linewidth]{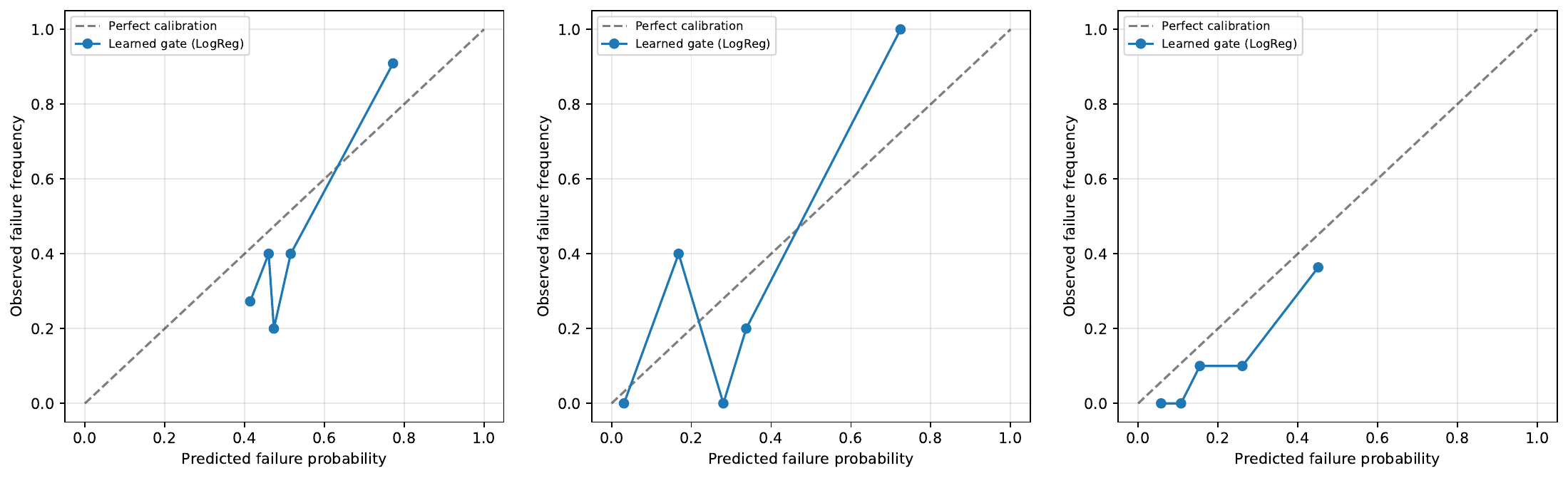}
\caption{Reliability diagrams for the learned logistic failure gate. From left to right, the panels show Qwen2.5-0.5B, Qwen2.5-1.5B, and Qwen2.5-3B. The dashed diagonal denotes perfect calibration. The finite calibration/test sample sizes make individual bins noisy, so ECE and Brier scores are reported alongside these diagrams rather than inferred visually.}
\label{fig:calibration_appendix}
\end{figure}

\begin{figure}[h]
\centering
\includegraphics[width=.72\linewidth]{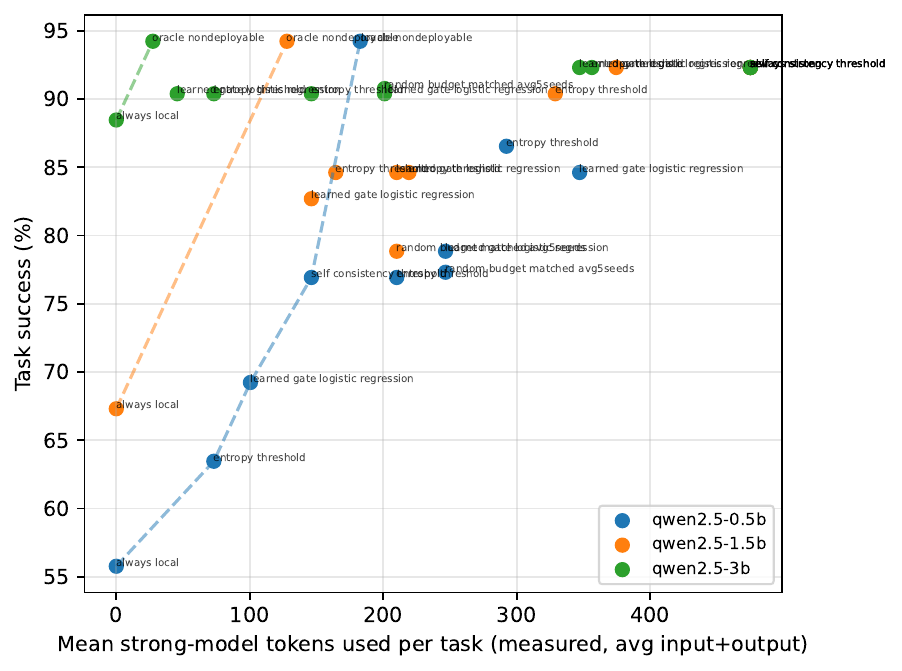}
\caption{Measured task success versus mean strong-model tokens per task (input+output). Each point is an executed routing configuration. The dashed connections are visual guides, not fitted performance models. The figure shows that token efficiency and task success must be considered jointly and that the preferred operating point depends on local model size.}
\label{fig:pareto_tokens}
\end{figure}

\begin{figure}[h]
\centering
\includegraphics[width=\linewidth]{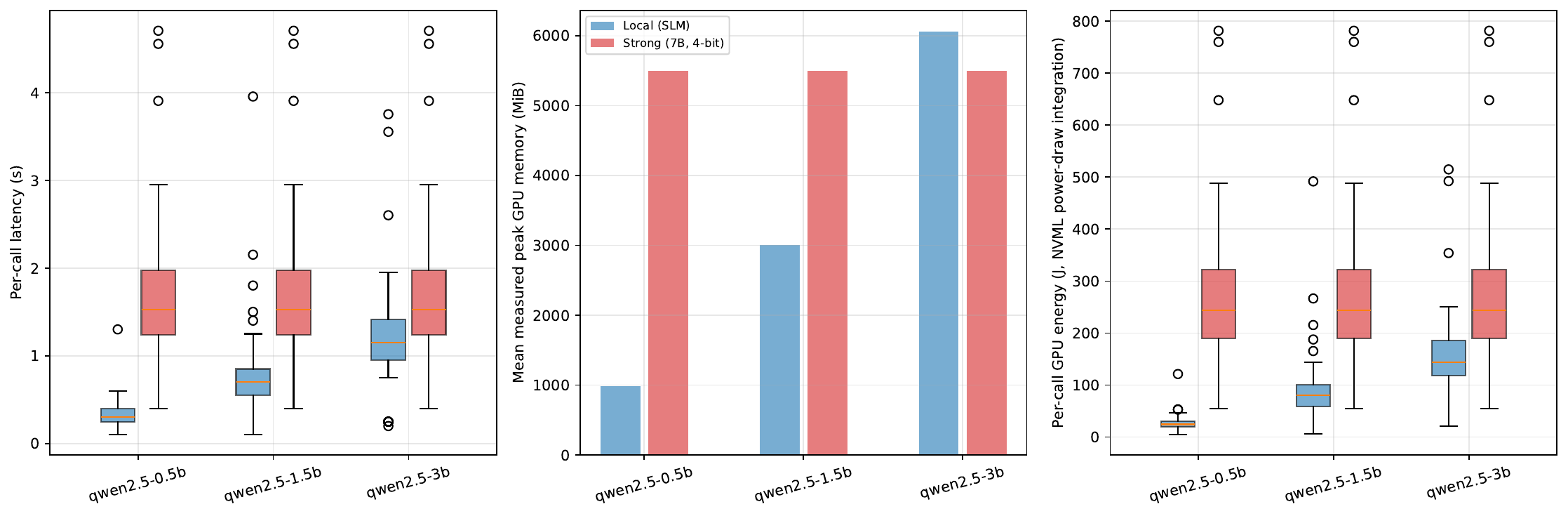}
\caption{Measured per-call latency, peak GPU memory, and NVML power-draw integration for the local SLMs and the shared 7B 4-bit strong model. These measurements are specific to the hardware/runtime in use. They are not hardware-independent efficiency claims.}
\label{fig:efficiency_measurements}
\end{figure}

\begin{figure}[h]
\centering
\includegraphics[width=\linewidth]{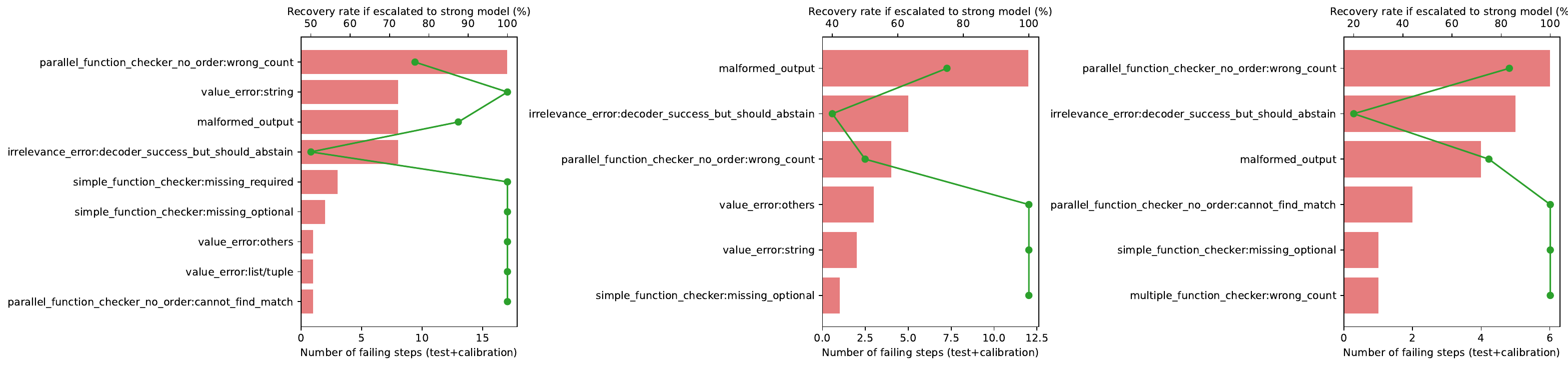}
\caption{Failure decomposition and escalation recoverability for the Qwen2.5-0.5B local model (left), Qwen2.5-1.5B local model (middle), and Qwen2.5-3B local model (right). The categories include irrelevance/abstention errors, malformed outputs, parallel-call count or matching errors, missing optional or required arguments, and value errors involving strings, lists/tuples, or other values. Red bars show observed local failure counts on the test and calibration sets, while the green line shows the recovery rate when the same failing step is escalated to the strong model. Recovery varies sharply by failure type, underscoring why local failure probability alone is not equivalent to the expected escalation benefit. Sparse categories should be interpreted cautiously.}
\label{fig:failure_breakdown}
\end{figure}

\begin{table}[t]
\centering\scriptsize
\caption{Measured efficiency statistics from the executed runs. Strong-model latency is repeated by row because the same Qwen2.5-7B 4-bit model is used.}
\resizebox{\linewidth}{!}{\begin{tabular}{l rr rr rr rr}
\toprule
SLM & Local in tok. & Local out tok. & Strong in tok. & Strong out tok. & Local lat. p50/p95 (s) & Strong lat. p50/p95 (s) & Local peak mem (MiB) & Local energy p50 (J) \\
\midrule
qwen2.5-0.5b & 427.9 & 41.1 & 427.9 & 47.1 & 0.301/0.576 & 1.528/3.385 & 990 & 24.31 \\
qwen2.5-1.5b & 427.9 & 47.9 & 427.9 & 47.1 & 0.702/1.638 & 1.528/3.385 & 3003 & 80.66 \\
qwen2.5-3b & 427.9 & 43.9 & 427.9 & 47.1 & 1.152/2.248 & 1.528/3.385 & 6058 & 143.86 \\
\bottomrule
\end{tabular}}
\label{tab:efficiency_appendix}
\end{table}
\paragraph{Calibration and efficiency details.} Across local-model sizes, the full logistic gate reaches AUROC 0.729/0.829/0.899 for 0.5B/1.5B/3B, respectively. Full-feature ECE is 0.154/0.212/0.135. Measured efficiency statistics are shown below; energy is reported as measured for this hardware only.

\subsection{Diagnostic plots from the executed runs}
The diagnostic plots make four complementary aspects of the routing behavior visible. First, Figure~\ref{fig:selective_risk} tests whether entropy is useful for ranking local actions by difficulty, independently of the learned gate. As the fraction of actions retained for local execution increases, the error rate among those actions also increases. The increase is steepest for the 0.5B model and shallowest for the 3B model, indicating that entropy-based selection separates easy from difficult actions more reliably for the weaker local model. This is a ranking result, however, and should not be interpreted as \method's success-escalation curve.

Second, Figure~\ref{fig:calibration_appendix} assesses whether the logistic gate's predicted failure probabilities are numerically meaningful, rather than merely correctly ordered. Curves close to the dashed diagonal indicate that actions assigned a probability near $p$ fail at roughly rate $p$ within that bin; systematic deviations show over- or under-confidence. The sparse bins should be read cautiously because the calibration and test sets are small. The corresponding ECE values of 0.154, 0.212, and 0.135 for the 0.5B, 1.5B, and 3B models, respectively, and the reported Brier scores provide the quantitative comparison.

Third, Figure~\ref{fig:pareto_tokens} shows the resource-quality trade-off in terms of mean strong-model input plus output tokens per task. Moving right means more remote inference, while moving upward means higher task success. The preferred point is therefore not necessarily the highest-success point: a configuration is attractive when it reaches a given success level with fewer strong-model tokens, and the best trade-off can differ by local-model size. The dashed lines only connect measured configurations. They are not interpolated or fitted performance curves. Finally, Figure~\ref{fig:efficiency_measurements} and Table~\ref{tab:efficiency_appendix} report the corresponding execution costs. They show how local-model size changes per-call latency and peak memory, and how those measurements compare with the shared 7B strong model. The NVML power integration is hardware-specific, not a portable energy estimate. Figure~\ref{fig:failure_breakdown} adds a failure-type view: recovery after escalation varies by category, so a high local failure probability alone does not imply the same expected benefit from cloud handoff for every action.

Together, these plots are descriptive diagnostics generated from the executed outputs rather than additional experiments. The selective-risk curve is entropy-ranked. The reliability diagrams use the learned logistic gate. The Pareto plot uses measured strong-model tokens per task, and the failure analysis combines test and calibration failures, as indicated on its axis. These distinctions keep the figures from being read as stronger evidence than the underlying runs support.

\section{Extended related work: UALA, RACER, and TokUR}
\label{app:extended-related}
This section expands the brief positioning given in the main text.

\paragraph{UALA \citep{han2024uala}.} The Uncertainty-Aware Language Agent integrates uncertainty estimation into the ReAct thought-action-observation loop for free-form question answering (HotpotQA, StrategyQA, MMLU). At each step, UALA scores the agent's current answer and uses an adaptive, calibration-set-derived threshold to decide whether to accept the answer, invoke an external tool (search), or, in a further extension, request human help. This is architecturally the closest prior work to \method's ``when to escalate'' decision, but the escalation target in UALA is a \emph{tool call} (or a human), not a stronger executor model, and the underlying action space is free-form text generation rather than structured function calling. \method{} instead assumes the tool has already been selected by the local model and asks which \emph{model} (local SLM or cloud LLM) should generate and execute the structured call, using schema- and argument-level features that have no analogue in UALA's free-form QA setting. The two ideas are complementary: an UALA-style gate could decide whether to call a tool at all and \method{} could then decide which executor handles that call.

\paragraph{RACER \citep{hao2026racer}.} RACER addresses single-turn LLM routing over a fixed candidate pool $\mathcal{M}$ with a known (or oracle-evaluable) ground-truth correctness set $G(x)$ per query. It reframes routing as the $\alpha$-Valid Optimal Routing problem: construct the smallest model \emph{set} $C(x)\subseteq\mathcal{M}$ such that the probability of excluding every correct model is at most a user-specified $\alpha$, using finite-sample conformal-style calibration (a monotone threshold over nonconformity scores, calibrated via Eq.~6 of \citet{hao2026racer}) to obtain a distribution-free guarantee under exchangeability. This differs from \method's setting in three ways that make a direct adaptation nontrivial: (i) RACER's calibration guarantee assumes i.i.d./exchangeable calibration and test \emph{queries}, whereas \method's decision points are \emph{steps within a trajectory}, which are not exchangeable with each other (later steps are conditioned on earlier routing decisions and tool outcomes); (ii) RACER assumes an oracle correctness label per query at calibration time, whereas our failure label for a step is itself counterfactual and can depend on the chosen executor's downstream behavior; and (iii) RACER selects a \emph{set} of models for aggregation, whereas \method{} must commit to exactly one executor per step under a cloud budget. A principled extension might calibrate at the trajectory level (treating whole trajectories, rather than individual steps, as the exchangeable unit) or use sequential/online conformal risk control; we view this as the most concrete next step toward closing the risk-control gap noted in Appendix~\ref{sec:limitations}.

\paragraph{TokUR \citep{zhang2026tokur}.} TokUR estimates token-level epistemic uncertainty for LLM chain-of-thought reasoning by introducing low-rank random weight perturbations during decoding, producing an ensemble of predictive distributions per token that are aggregated into a sequence-level uncertainty signal correlated with answer correctness on mathematical reasoning benchmarks. Our \emph{intrinsic} feature family (token NLL, entropy, tool-choice margin; Appendix~\ref{app:multiturn-setup} and Table~\ref{tab:features_appendix}) targets a similar goal but uses only plain decode-time statistics from a single forward pass, without weight perturbation or ensembling, since our SLMs must run with tight per-step latency budgets (Section~4, ``Efficiency''). Incorporating a lightweight, TokUR-style perturbation ensemble into the intrinsic feature family, traded off against its added local latency, is a natural direction for improving the token-uncertainty signal used by \method's gate.

\section{Limitations and broader impact}
\label{sec:limitations}

The evaluation remains narrower than the full deployment setting in Figure~\ref{fig:arch}. The multi-turn experiment tests re-routing across actions and reports a 21-point trajectory-success improvement over local-only at 30\% cloud actions.  {Its comparisons are accompanied by paired bootstrap intervals and McNemar tests (Section~\ref{sec:experiments}), but they rest on 100 scripted tasks, and the differences among \method, random, query-level, and self-consistency routing are small relative to this sample size.} The original single-step test has only 52 held-out tasks and correspondingly wide confidence intervals {, and its baselines have unequal realized escalation rates (Appendix~\ref{app:budget}). All baselines escalate more than \method, and rate-matched variants are given in Table~\ref{tab:budget}}. The evaluation also covers one model family, one strong model (Qwen2.5-7B 4-bit), and an in-distribution benchmark configuration, so robustness to other strong backends, distribution shift, and other agent benchmarks (e.g., ToolSandbox, WebArena) remains outside the empirical scope.  {Results are not stratified by risk tier or task difficulty, although the tags exist (Appendix~\ref{app:multiturn-setup}).} Finally, the learned feature set is not uniformly beneficial: removing risk metadata improves the 1.5B logistic gate  {(at a higher escalation rate; the equal-budget comparison is in Appendix~\ref{app:budget})}, and the 3B router is not clearly better than random routing; we read this as feature interaction and possible overfitting on a heterogeneous, modestly sized feature set rather than evidence against any one family.  {We therefore do not claim that the risk features improve task success.}

Two further limitations bear on deployment readiness. First, \method's gate is calibrated empirically and gives no finite-sample risk-control guarantee: unlike RACER's $\alpha$-VOR routing, which provably bounds misrouting probability \citep{hao2026racer} (Appendix~\ref{app:conformal-background}), our thresholds can be violated on a given sample; adapting such a guarantee to a stateful, step-level setting is open work. Second, local self-correction is explored only lightly (sample-based consistency plus one bounded retry), not deeper deliberation, so how much of the local/strong gap more local compute could close remains unknown.

The risk taxonomy is a research annotation, not an authorization policy. Appendix~\ref{app:risk-taxonomy} gives worked examples mapping representative tool calls to each tier. ``Wrong risky'' measures errors on actions labeled R2 or above in this benchmark. It is not a real-world harm rate and does not constitute a safety guarantee. Likewise, remote-token count is only a disclosure-volume proxy, not a privacy guarantee: the minimized packet still transmits task content, and we do not evaluate leakage or adversarial settings. Energy and latency depend on hardware and runtime. Strong-model escalation can fail, as shown by the 92.3\% strong-only success rate, and correlated local/strong failures remain possible. The broader-impact objective is to reduce unnecessary remote inference while retaining reliability, but high-impact actions still require deterministic application policy and, where appropriate, human authorization.

\color{black}
\end{document}